\pdfoutput=1
\documentclass[11pt,a4paper,table, dvipsnames]{article}
\usepackage[table,dvipsnames]{xcolor}
\usepackage[hyperref]{emnlp2020}
\usepackage{times}
\usepackage{latexsym}

\usepackage{multirow}
\usepackage{amsmath}
\usepackage{multicol}
\usepackage{subcaption}
\usepackage{url}
\usepackage{longtable}
\usepackage{latexsym}
\usepackage{enumitem}
\usepackage{comment}
\usepackage{times}
\usepackage{multirow}
\usepackage{array}
\usepackage{latexsym}
\usepackage{amssymb}
\usepackage{tabularx}
\usepackage{booktabs}
\usepackage{graphicx}
\usepackage{url}
\usepackage{graphics}
\usepackage{lipsum}
\usepackage{enumitem}
\usepackage{rotating}
\usepackage{array}
\usepackage{pgf}

\newcolumntype{P}[1]{>{\centering\arraybackslash}p{#1}}

\usepackage{microtype}

\aclfinalcopy 

\title{
Natural Language Processing
for Achieving Sustainable Development:\\
the Case of
Neural Labelling to Enhance
Community Profiling
}

\author{
  Costanza Conforti$^{1,2}$,
  Stephanie Hirmer$^{1,3}$
  David Morgan$^{4}$,
  Marco Basaldella$^{2}$,
  Yau Ben Or$^{1}$\\
  $^1$Rural Senses Ltd.\\
  $^2$Language Technology Lab, University of Cambridge \\
  $^3$Energy and Power Group, University of Oxford\\
  $^4$Centre for Sustainable Development, University of Cambridge \\
  {\tt info@ruralsenses.com}
  }

\date{}

\begin{document}
\maketitle
\begin{abstract}

In recent years, there has been an increasing interest in the application of Artificial Intelligence – and especially Machine Learning – to the field of Sustainable Development (SD).
However, until now, NLP has not been systematically applied in this context.
In this paper, we show the high potential of NLP to enhance project sustainability.
In particular, we focus on
the case of 
community profiling in developing countries,
where, in contrast to the developed world, a notable data gap exists.
Here, NLP could help to address the cost and time barrier of structuring qualitative data that prohibits its widespread use and associated benefits.
%
We propose the new extreme multi-class multi-label \textit{Automatic User-Perceived Value classification} task. 
%
We release \textit{Stories2Insights} (S2I), an expert-annotated dataset of interviews carried out in Uganda, we provide a detailed corpus analysis, and we implement a number of strong neural baselines to address the task.
Experimental results show that the problem is challenging, and leaves considerable room for future research at the intersection of NLP and SD.
%

%
%

\end{abstract}

\section{Introduction}

Sustainable Development (SD) 
is an interdisciplinary field which studies the integration and balancing of economic, environmental and social concerns to tackle the broad goal of achieving inclusive and sustainable growth~\cite{brundtland1987our,keeble1988brundtland,sachs2015age}. 
As a collective, trans-national effort toward sustainability, in 2015 the United Nations approved the \textit{2030 Agenda}~\cite{united2015transforming}, which identifies 17 Sustainable Development Goals (SDGs) to be reached by 2030~\cite{lee2016transforming}. In recent years, there has been increasing recognition of the fundamental role played by data in achieving the objectives set out in the SDGs~\cite{griggs2013policy,nilsson2016policy,vinuesa2020role}.

In this paper, we focus on data-driven planning and delivery of 
projects\footnote{Examples of projects for SD include
\textit{physical infrastructures} (as the installation of a solar mini-grid to provide light~\cite{BHATTACHARYYA2012260}) or of \textit{programmes} to change a population's behaviour
(as the awareness raising campaigns against HIV transmission implemented by~\newcite{avert}).}  
which address one or more of the SDGs in a developing country context.
%
%
When dealing 
with developing countries, 
a deep understanding of project beneficiaries' 
needs and values (hereafter referred to as \textit{User-Perceived Values} or UPVs, \citet{HIRMER2016UPVmethod})
is of particular importance.
This is because 
beneficiaries 
with limited financial means 
are especially good at assessing needs and values
~\cite{hirji2015accelerating}.
When a project fails to create value to a benefiting community, the community is less likely to care about its continued operation~\cite{watkins2012guide,chandler2013aspirations,hirmerthesis} and
as a consequence, the chances of the project's long-term success is jeopardised~\cite{bishop2010marketing}.
Therefore,
comprehensive community profiling\footnote{\textit{Community profiling}
is the
detailed and holistic description of a community's needs and resources~\cite{blackshaw2010key}.}
plays a 
key role in 
understanding what is important for a community and act upon it,
thus ensuring a project's sustainability~\cite{van2019community}.

%
Obtaining data with such characteristics
requires knowledge extraction from qualitative interviews
which come in the form of unstructured free text
~\cite{DBLP:conf/lrec/SaggionSMS10,DBLP:conf/icacci/ParmarMDP18}.
This step is usually done manually by domain experts~\cite{lundegaard2007conflicts}, which further raises the costs.
Thus, structured qualitative data is often unaffordable
for project developers.
As a consequence, project planning
heavily relies upon sub-optimal
aggregated statistical data,
like
household surveys
{~\cite{world2016world}}
or 
remotely-sensed satellite imagery
~\cite{bello2014satellite,jean2016combining}, which unfortunately is of considerable lower resolution in developing countries.
Whilst these quantitative data sets are important and necessary, they are insufficient to ensure successful project design, lacking insights on UPVs that are crucial to success.
%
In this context, the application of NLP techniques can help to make qualitative data more accessible to project developers by dramatically reducing time and costs to structure data.
However,
despite 
having been successfully applied to many other
domains – ranging from biomedicine~\cite{simpson2012biomedical}, to law~\cite{kanapala2019text} and finance~\cite{loughran2016textual} – to our knowledge, NLP
has not yet been applied to the field of SD 
{in a systematic and academically rigorous format}\footnote{We have found sporadic examples of the application of NLP, e.g.~for analysing data from a gaming app used in a developing country~\cite{pulsenlp}.
}.

In this paper, we make the following contributions:
\textsc{(1)}~we articulate the potential of NLP to enhance SD—at the time of writing this is the first time NLP is systematically applied to this field; 
%
%
%
%
\textsc{(2)}~as a case-study
at the intersection between NLP and SD,
we focus on enhancing project planning in the context of a developing country, namely Uganda;
%
%
\textsc{(3)}~we propose the new task of 
\textit{UPV Classification},
which consists in labeling
qualitative interviews
{using an annotation schema developed in the field of SD};
\textsc{(4)}~we annotate and release 
\textit{Stories2Insights}, a corpus of UPV-annotated interviews in English;
\textsc{(5)}~we provide a set of strong neural baselines for future reference; and 
\textsc{(6)}~{we show – through a detailed error analysis – that the task is challenging and important, and we hope it will raise interest from the NLP community.}

\section{Background}

\subsection{Artificial Intelligence for Sustainable Development }
While NLP has not yet been applied to the field of SD, in recent years there have been notable applications of Artificial Intelligence~(AI) in this area.
This is testified by
%
the rise of young research fields that seek to help meet the SDGs, as \textit{Computational Sustainability}~\cite{gomes2019computational} and
\textit{AI for Social Good}~\cite{hager2019artificial,DBLP:journals/corr/abs-2001-01818}.

In this context, 
Machine Learning, in particular in the field of Computer Vision~\cite{de2018machine}, has been applied
to contexts ranging from
conservation biology~\cite{kwok2019ai}, to poverty~\cite{blumenstock2015predicting} and
slavery mapping~\cite{DBLP:journals/remotesensing/FoodyLBLW19}, to deforestation and water quality monitoring~\cite{DBLP:journals/remotesensing/HollowayM18}.

\subsection{Ethics of AI for Social Good }
Despite its positive impact,
it is important to recognise that some AI techniques 
can act both as an enhancer and inhibitor of sustainability. As recently shown by~\newcite{vinuesa2020role},
AI might inhibit meeting a considerable number of
targets across the SDGs and may result in inequalities within and across countries due to application biases.
Understanding the implications of AI and its related fields on SD, or Social Good more generally, is particularly important for countries where action on SDGs is being focused and 
where issues are most acute~\cite{unescoai,unescolearningweek}.
%
%

\subsection{Project biases }
Various works highlight the importance of understanding the local context and engaging with local stakeholders, including beneficiaries, to achieve project sustainability. 
Where such information is not available, 
projects are designed and delivered 
based on the judgment of other actors (e.g. project funders, developers or domain experts, \cite{risal2014mismatch, axinn1988international, harman2014international}).
Their judgment
, in turn, is subject to biases \cite{kahneman2011thinking} that are shaped by past experiences, beliefs, preferences 
and worldviews:
such biases can include, for example,
%
preferences towards a specific sector (e.g.~energy or water), technology (e.g.~solar, hydro) or gender-group (e.g.~solutions which benefit a gender disproportionately), which are 
pushed without considering the local needs.

NLP has the potential to increase the availability of 
community-specific data to key decision makers and ensure project design is properly informed and appropriately targeted.
However, careful attention needs to be paid to the potential for bias in data collection resulting from the interviewers \cite{bryman2016social}, {as well as the potential to introduce new bias through NLP.}

\section{User-Perceived Values (UPVs) for Data-driven Sustainable Projects}
\label{sec:upv_theory}

\begin{figure*}[t!]
 \begin{subfigure}[b]{0.28\textwidth}
 \centering
    \includegraphics[width=4.2cm]{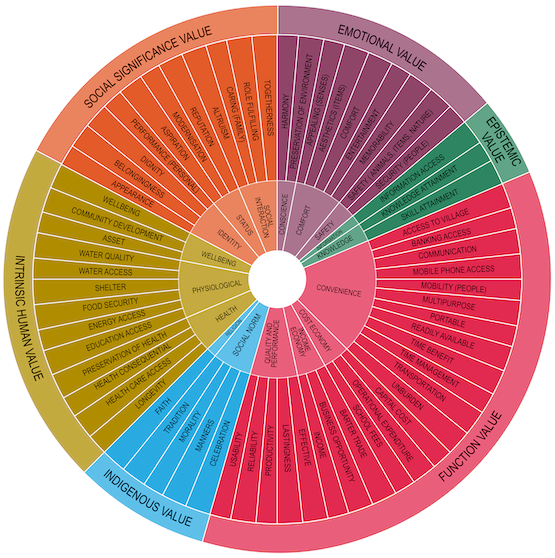}
    \caption{User-Perceived Value wheel.}
    \label{fig:upv_wheel}
 \end{subfigure}
  \begin{subfigure}[b]{0.7\textwidth}
  \centering
    \includegraphics[width=10.2cm]{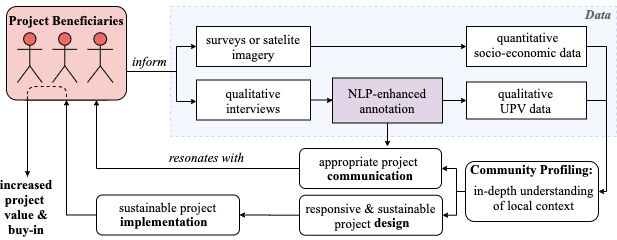}
    \caption{Flowchart of the intersection between NLP (purple square) and the delivery of SD projects.}
    \label{fig:proj_dev_schema}
 \end{subfigure}
 \caption{Using UPVs (\ref{fig:upv_wheel}) to build sustainable projects: note 
 the role of NLP (purple square in~\ref{fig:proj_dev_schema}).}
\end{figure*}

\subsection{The User-Perceived Values (UPV) Framework.}
As a means to obtain qualitative data with the characteristics mentioned above,
we adapt the User-Perceived Values (UPV) framework~\cite{hirmerthesis}.
The UPV framework builds on value theory, which is widely used in marketing and product design in the developed world~\cite{sheth1991we,woo1992cognition,solomon2002value,boztepe2007user}.
Value theory assumes
that a deep connection exists between
what consumers perceive as important
and
their inclinations to adopt a new product or service~\cite{nurkka2009capturing}.

In the context of developing countries,
our UPV framework identifies a set of 58 UPVs
which can be used to
frame the wide range of perspectives on what is of greatest concern to project beneficiaries~\cite{HIRMER2016UPVmethod}.
UPVs {(or \textit{tier 3} (T3) values)} can be clustered into 17 \textit{tier 2} (T2) value groups, each one embracing a set of similar T3 values; in turn, T2 values can be categorized into 6 \textit{tier 1} (T1) high-level value pillars, as follows:~\cite{HIRMER2014145}:
\begin{enumerate}[noitemsep,topsep=0pt,leftmargin=*]
\item \textit{Emotional}: contains the T2 values \textit{Conscience},
\textit{Contentment},
\textit{Human Welfare} (tot.~9 T3 values)
\item \textit{Epistemic}:
contains the T2 values
\textit{Information} and \textit{Knowledge} (tot.~2 T3 values)
\item \textit{Functional}:
contains the T2 values
\textit{Convenience}, 
\textit{Cost Economy}, 
\textit{Income Economy} and
\textit{Quality and Performance} (tot.~21 T3 values)
\item \textit{Indigenous}:
containing the T2 values
\textit{Social Norm} and
\textit{Religion} (tot.~5 T3 values)
\item \textit{Intrinsic Human}: \textit{Health},
\textit{Physiological} and
\textit{Quality of Life} (tot.~11 T3 values)
\item \textit{Social significance}:
contains the T2
\textit{Identity},
\textit{Status} and 
\textit{Social Interaction} (tot.~11 T3 values)
\end{enumerate}
\noindent
The interplay between T1, T2 and T3 values is graphically depicted in the \textit{UPV Wheel} (Figure~\ref{fig:upv_wheel}).
See Appendix A for the full set of UPV definitions.

\subsection{Integrating UPVs into Sustainable Project Planning. }
%
The UPV approach 
offers a theoretical framework to 
place communities at the centre of project design
(Figure~\ref{fig:proj_dev_schema}).
Notably, it allows to~(a)~facilitate more responsible and beneficial project planning~\cite{gallarza2006value};
and ~(b)~enable effective communication with rural dwellers.
The latter allows the use of messaging of project benefits in a way that resonates with the beneficiaries' own understanding of benefits, as discussed by~\newcite{hirji2015accelerating}.
This results in a higher end-user acceptance, because the initiative is perceived to have personal value to the beneficiaries:
as a consequence, community commitment will be increased,
eventually enhancing the project success rate and leading to more sustainable results \cite
{hirmerthesis}.

\begin{figure*}[t!]
 \begin{subfigure}[b]{0.3\textwidth}
    \centering
    \includegraphics[height=4.2cm]{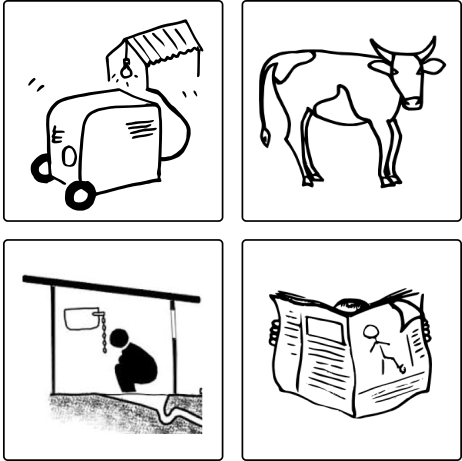}
    \caption{}
    \label{fig:cards}
 \end{subfigure}
 \hfill
 \begin{subfigure}[b]{0.35\textwidth}
    \centering
    \includegraphics[height=4.2cm]{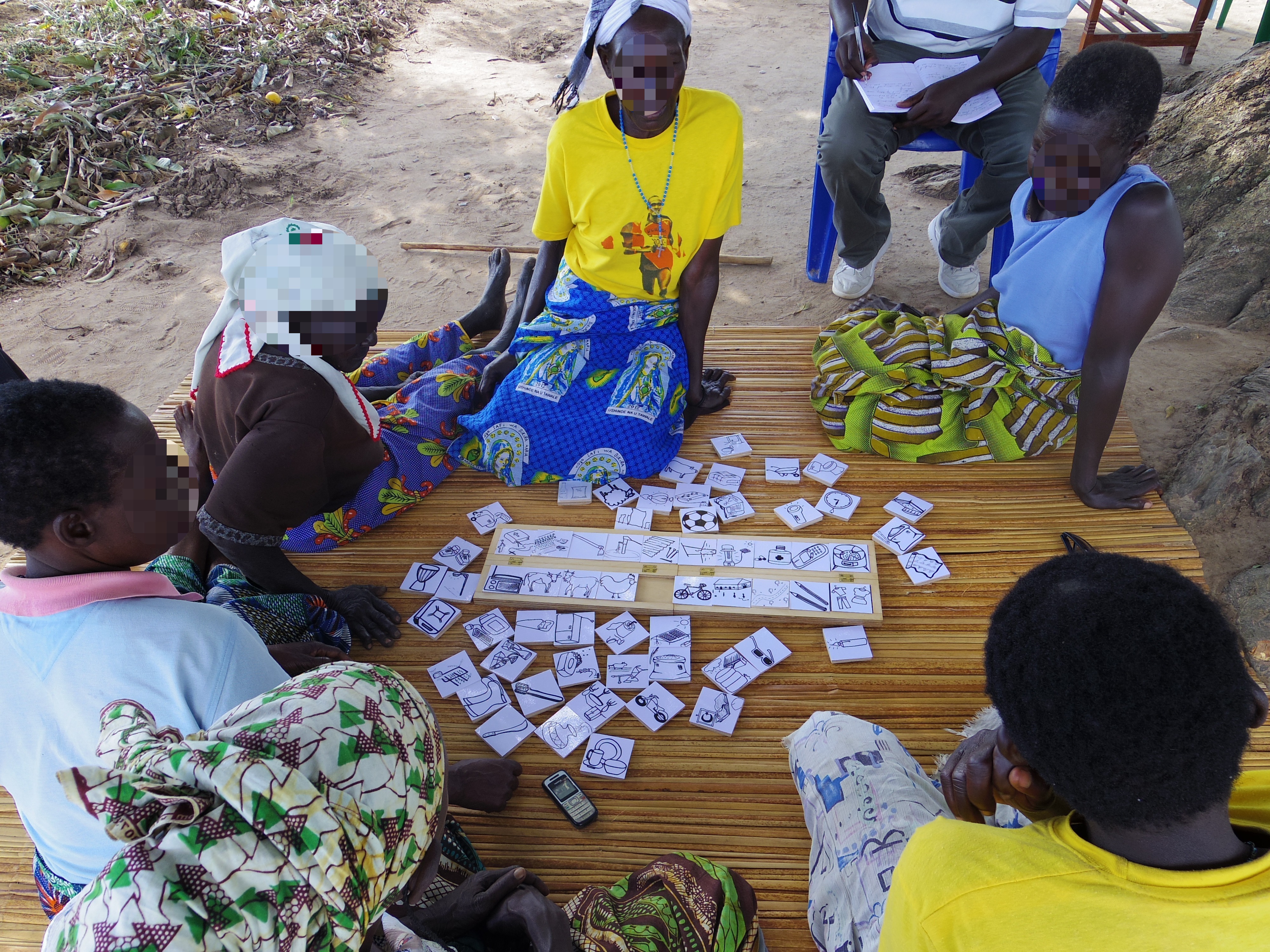}
    \caption{}
    \label{fig:game}
 \end{subfigure}
 \hfill
 \begin{subfigure}[b]{0.25\textwidth}
 \centering
 \includegraphics[height=4.1cm]{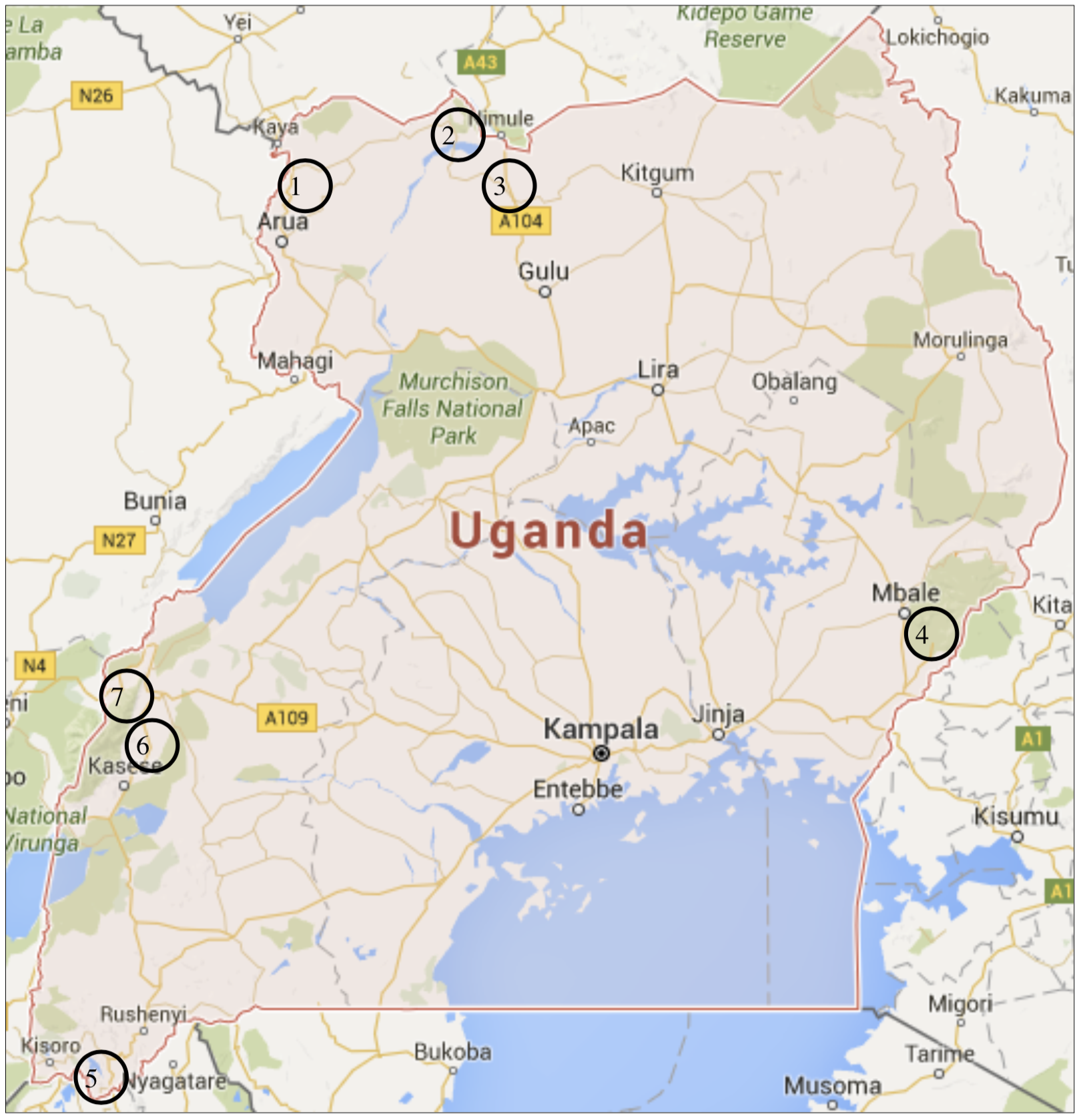}
    \caption{}
    \label{fig:map}
 \end{subfigure}
 \caption{Playing the UPV game in Uganda.
 From left to right:
 \ref{fig:cards})~Cards for the items \textit{generator}, \textit{cow}, \textit{flush toilet} and \textit{newspapers} (adapted to the Ugandan context with the support of international experts and academics from the U.~of Cambridge;
 \ref{fig:game})~Women playing the UPV game in village \textsc{(1)}\footnotemark;
 \ref{fig:map})~Map of case-study villages.}
\end{figure*}
\subsection{The role of NLP to enhance Sustainable Project Planning. }
%
Data conveying the beneficiaries' perspective is seldom
considered in practical application,
%
mainly due to the fact that
it comes in the form of unstructured qualitative interviews.
As introduced above, data needs to be \textit{structured} in order to be 
useful~\cite{OECD,unstats}.
This makes
the entire process very long and costly, thus making it almost prohibitive to afford in practice for most small-scale projects.
In this context,
the role of AI, and more specifically NLP,
can have a yet unexplored opportunity.
Implementing successful NLP systems to 
automatically perform the annotation process on interviews (Figure~\ref{fig:proj_dev_schema},
purple square),
which constitutes the major bottleneck
in the project planning pipeline (Section~\ref{sec:corpus}),
would dramatically speed up
the entire project life-cycle  and drastically reduce its costs.

In this context, we introduce the task of \textit{Automatic UPV classification},
which consists of
annotating each sentence of a given input interview with the appropriate UPV labels which are (implicitly) conveyed
by the interviewee.

\section{The \textit{Stories2Insights} Corpus: a Corpus Annotated for User-Perceived Values}
\label{sec:corpus_all}

To enable research in UPV classification,
we release S2I, a corpus of labelled reports
from 7 rural villages in Uganda (Figure~\ref{fig:map}).
In this Section,
we report on the corpus collection and annotation procedures and outline the challenges this poses for NLP.

\noindent
\subsection{Building a Corpus with the UPV game}
\label{sec:corpus}

\noindent
\textbf{The UPV game. }
As widely recognised in marketing practice~\cite{van2005consumer},
consumers are usually unable to articulate
their own values and needs~\cite{ulwick2002turn}. This requires the use of methods that elicit what is important, such as laddering \cite{reynolds2001laddering} or Zaltman Metaphor Elicitation Technique (ZMET) \cite{coulter2001interpreting}.
To avoid direct inquiry~\cite{pinegar2006customers},
\newcite{HIRMER2016UPVmethod}
developed an approach
to identify perceived values
in low-income settings by means of a game (hereafter referred to as 
\textit{UPV game}).
Expanding on the items proposed by~\newcite{PeaceChildInternational},
the UPV game makes reference to
46 everyday-use items 
in rural areas\footnote{Such items included livestock (\textit{cow, chicken}), basic electronic gadgets (\textit{mobile
phone, radio}), household goods (\textit{dishes, blanket}), and
horticultural items (\textit{plough, hoe})~\cite{hirmerthesis}.},
which are
graphically depicted~(Figure~\ref{fig:cards}).
The decision to represent items graphically stems from the high level of illiteracy across developing countries~\cite{unesco2013adult}.

Building on the techniques proposed by 
Coulter \textit{et al.} ~\shortcite{coulter2001interpreting} and Reynolds \textit{et al.} \shortcite{reynolds2001laddering},
the UPV game
is framed in the form of semi-structured interviews:\\\noindent
\textsc{(1)}~participants are asked to select 20 
items, based on what is most important to them (\textit{Select stimuli}),\\\noindent
\textsc{(2)}~
to rank them in order of 
importance;
and finally,\\\noindent
\textsc{(3)}~they have to give reasons as to why an item was important to them. \textit{Why-probing} was used to encourage discussion (\textit{Storytelling}).
\footnotetext{While permission of photographing was granted from the participants, photos were pixelised to protect their identity.}

\noindent
\textbf{Case-Study Villages.~}
7 rural villages were studied: 
3 in the West Nile Region (Northern Uganda);
1 in Mount Elgon (Eastern Uganda);
2 in the Ruwenzori Mountains (Western Uganda);
and 1 in South Western Uganda.
All villages are located in remote areas far from the main roads (Figure~\ref{fig:map}).
{A total of 7 languages are spoken across the villages\footnote{Rukonjo, Rukiga, Lugwere and Swahili (Bantu family);
Sebei/Sabaot, Kupsabiny, Lugbara (Nilo-Saharan family).}.}

\begin{figure*}[t!]
    \centering
    \includegraphics[width=0.9\textwidth]{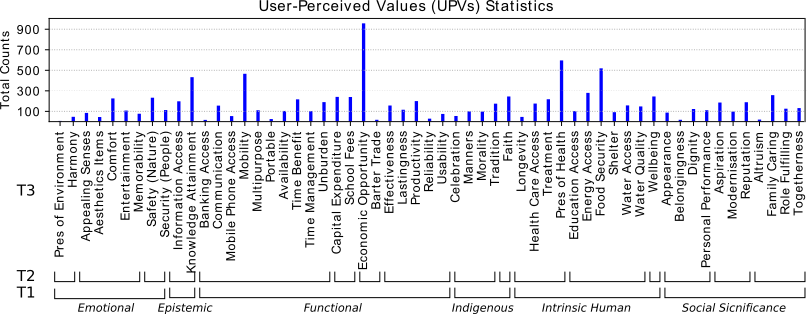}
    \caption{UPV frequencies from the S2I corpus 
    (see Appendix A for UPV definitions).
    }
    \label{fig:stats_aggregated}
\end{figure*}

\noindent
\textbf{Data Collection Setting and Guidelines for Interviewers.~}
%
For each village, 3 native speaker interviewers 
guided the UPV game. 
To ensure consistency and data quality, 
a two-day training workshop was held at Makerere University
(Kampala, Uganda), and a local research assistant oversaw the entire data collection process 
in the field.

\noindent
\textbf{Data Collection.~}
12 people per village were interviewed, 
consisting of an equal split between men and women with varying backgrounds and ages. 
In order to gather complete insight into the underlying decision-making process –
which might be influenced by the context~\cite{barry2008determining} –
interviews were conducted both individually and in groups of 6 people following standard focus group methods~\cite{silverman2013doing,bryman2016social}.
Each interview lasted around
90 minutes.
The data collection process took place over a period of 3~months and resulted in a total of 119 interviews.

\noindent
\textbf{Ethical Considerations.~}
Participants received 
compensation in the amount of 
1 day of labour.
%
An informed consent form 
was read out loud by the interviewer prior to the UPV game, to cater for the high-level of illiteracy amongst participants.
To ensure 
integrity, a risk assessment following the University of Cambridge's
\textit{Policy on the Ethics of Research Involving Human Participants and Personal Data} 
was completed.
To protect the participants' identity, locations and proper names were anonymized.

\noindent
\textbf{Data Annotation. }
The interviews were translated\footnote{Note that translating into English (or other languages commonly spoken in international workplaces, \url{https://www.un.org/en/sections/about-un/official-languages/}) is often a crucial step when applying knowledge to practical application in SD, in this case project decision-making~\cite{bergstrom2012knowledge}.
}
into English, analysed and annotated 
by domain experts\footnote{A team of researchers from the Department of
Engineering for Sustainable Development, 
supported by researchers in Development Studies and Linguistics, all at 
the University of Cambridge.}
using the
computer-assisted 
qualitative data analysis software
\textit{HyperResearch}~\cite{hesse1991hyperresearch}. 
To ensure consistency across interviews, 
they were annotated following
~\newcite{bryman2012mixed}, 
using cross-sectional indexing
\cite{mason2002organizing}.
Due to the considerable size of collected data, the annotation process took around 6 months. 

\subsection{Corpus Statistics and NLP Challenges}
\label{sec:corpus_nlp}

\noindent
We obtain a final corpus of 
5102
annotated utterances from the interviews. 
Samples 
present an average length of 
20 tokens.
The average number 
of samples per T3 label is 
169.1, 
with an extremely skewed distribution:
the most frequent T3, \textit{Economic Opportunity}, occurs 
957 times, while the least common, \textit{Preservation of the Environment}, only {7} (Figure~\ref{fig:stats_aggregated}). 

58.8\% of the samples are associated with more than 1 UPV, and 22.3\% with more than 2 UPVs (refer to Appendix B for further details on UPV correlation).
Such characteristics make 
UPV classification 
highly challenging to model:
the task is 
an extreme multi-class multi-label problem, with high class imbalancy. 
Imbalanced label distributions 
pose a challenge for many NLP applications – as sentiment analysis~\cite{li2011imbalanced}, sarcasm detection~\cite{liusarcasm}, and NER~\cite{tomanek2009reducing} – but 
are not uncommon in user-generated data~\cite{imran2016twitter}.
%
The following interview 
excerpt illustrates the multi-class multi-label characteristics of the problem: 
\begin{enumerate}[noitemsep,topsep=0pt,leftmargin=*]
\item \textit{If I have a flush
toilet in my house 
I can be a king of all kings
because I can’t go out on those squatting latrines} [Reputation][Aspiration]
\item \textit{And recently I was almost {rapped}} (sic.) \textit{
when I escorted my son to the latrine 
} [Security]
\item \textit{That [...] we have so many cases in our village of kids that fall into pit latrine} [Safety][Caring]
\end{enumerate}
%
\begin{figure*}
    \centering
    \begin{minipage}{0.70\textwidth}
        \centering
        \includegraphics[width=0.95\textwidth]{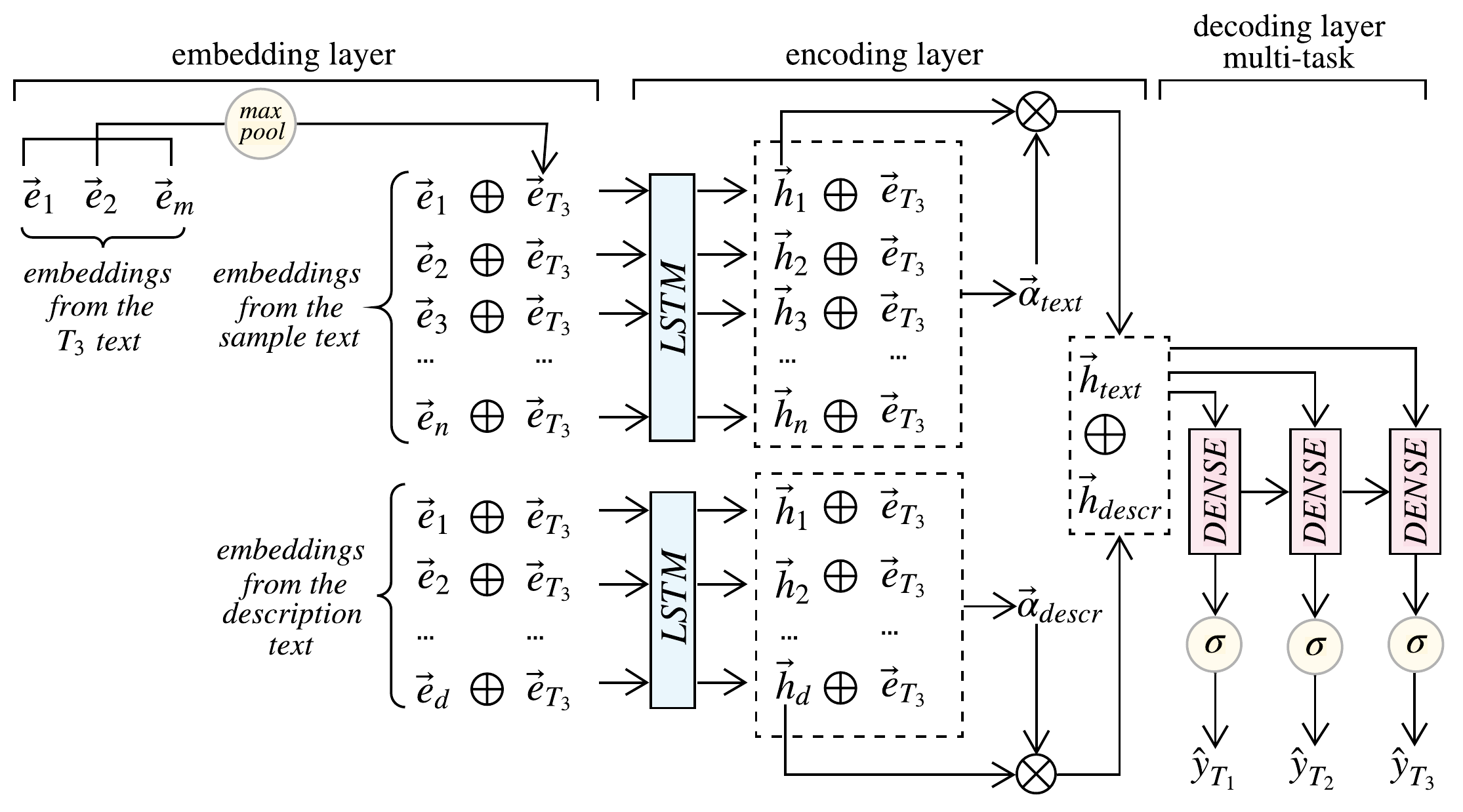} 
        \caption{Multi-task neural architecture for UPV classification.}
        \label{fig:architecture}
    \end{minipage}\hfill
    \begin{minipage}{0.27\textwidth}
        \centering
        \includegraphics[width=\textwidth]{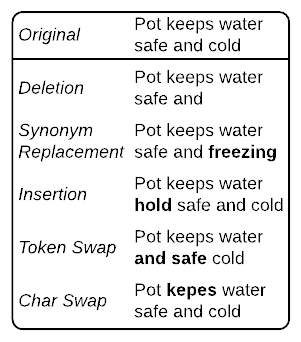} 
        \caption{Examples of negative samples generated through data augmentation.}
        \label{fig:augm}
    \end{minipage}
\end{figure*}

\noindent
Further challenges for NLP are introduced by
the frequent use of non-standard grammar
and poor sentence structuring,
which often occur in oral production~\cite{cole1995challenge}.
Moreover, manual transcription of interviews
may lead to spelling errors, thus increasing OOVs.
This is illustrated in the below excerpts (spelling errors are underlined):
\begin{itemize}[noitemsep,topsep=0pt,leftmargin=*]
    \item \textit{Also men like phone \underline{there} are so jealous for their women for example like in the morning my husband called me and asked that are you in church; so that's why they picked a phone.
    }
    \item \textit{A house keeps secrecy for example 
    [...]
    I can be bitten by a snake if I had sex outside [...] you see, 
    me I cannot because \underline{may} child is looking for \underline{mangoes} in the bush and finds me there, 
    how do I explain, can you imagine!!}
\end{itemize}
\noindent

\section{User-Perceived Values Classification}

As outlined above, given an input interview,
the task consists in annotating each sentence with the appropriate UPV(s).
The extreme multi-class multi-label quality of the task (Section~\ref{sec:corpus_nlp})
makes it impractical
to tackle as a standard \textit{multi-class classification} problem—where,
given an input sample $x$, 
a system is trained to predict its label from a tagset $T=\{l_1, l_2, l_3\}$ as~$x\rightarrow l_2$ (i.e.~[0,1,0]).
Instead, 
we model the task as a \textit{binary classification} problem:
given 
$x$, 
the system learns to predict its \textit{relatedness} with each one of the possible labels, i.e.~$(x, l_1) \rightarrow 0$,
$(x, l_2) \rightarrow 1$
and
$(x, l_3) \rightarrow 0$
\footnote{Note that this is different to the classic 
\textit{binary relevance} method, where a \textit{separated} binary classifier is learned for each considered label~\cite{DBLP:journals/ml/ReadPHF11}.}.

We consider the samples from the S2I corpus as \textit{positive instances}.
Then, we generate three kinds of \textit{negative instances}
by pairing the sample text with random labels.
To illustrate,
consider the three T2 classes \textit{Convenience}, \textit{Identity} and \textit{Status}, which contain the following T3 values:
\begin{itemize}[topsep=0pt,noitemsep,leftmargin=*]
\item 
\textit{Contentment}$_{T2}$ = \{\textit{Aesthetic}$_{T3}$, \textit{Comfort}$_{T3}$, ...\}
 \item \textit{Identity}$_{T2}$ = \{\textit{Appearance}$_{T3}$, 
 \textit{Dignity}$_{T3}$...\}
 \item \textit{Status}$_{T2}$ = \{\textit{Aspiration}$_{T3}$, \textit{Reputation}$_{T3}$, ...\}
\end{itemize}
Moreover, 
\textit{Contentment}$_{T2}$ $\in$ \textit{Emotional}$_{T1}$ and \{\textit{Identity}$_{T2}$, \textit{Status}$_{T2}$\}
$\in$ \textit{SocialSignificance}$_{T1}$. 
Given a sample $x$ and its gold label \textit{Aspiration}$_{T3}$, we can generate the following training samples:
\begin{itemize}[topsep=0pt,noitemsep,leftmargin=*]
    \item $(x, \text{\textit{Aspiration}}_{T3})$ is a \textit{positive sample};
    \item $(x, \text{\textit{Reputation}}_{T3})$ is a \textit{mildly negative sample}, as $x$ is linked with a wrong T3 with the same T2;
    \item $(x, \text{\textit{Dignity}}_{T3})$ is \textit{negative sample}, as $x$ is a associated with a wrong T3 from a different T2 class, but both T2 classes belong to the same T1; and
    \item $(x, \text{\textit{Aesthetic}}_{T3})$ is a \textit{strictly negative sample}, as $x$ is associated with a wrong label from the another T2 class in a different T1.
\end{itemize}
In this way, during training the system is exposed to positive (real) samples and negative (randomly generated) samples. 

A UPV classification system should satisfy the following desiderata:
(1)~it should be relatively light, given that it will be used in the context of developing countries, which may suffer from access bias\footnote{
With \textit{access bias}  we refer to contexts with
limited computational capacity and cloud services accessibility.}
and
{(2)~the goal of such a system
isn't to completely replace the work of human SD experts,
but rather to
reduce the time needed for interview annotation.
In this context, false positives are quick to notice and delete, while false negatives are more difficult to spot and correct.
Moreover, when assessing a community's needs and values, missing a relevant UPV is worse than including one which wasn't originally present.
For these reasons, recall is particularly important for a UPV classifier.}
%

In the next Section, we provide a set of strong baselines for future reference.

\subsection{Neural Models for UPV Classification}

\subsubsection{Baseline Architecture}

\noindent
\textit{Embedding Layer. }
The system receives an input sample $(x, T3)$, where $x$ is the sample text $(e_1, ..., e_n)$,
$T3$ is the T3 label
as the sequence of its tokens
$(e_1, ..., e_m)$,
and $e_i$ is the word embedding representation of a token at position $i$.
We obtain a T3 embedding $e_{T3}$ for each T3 label using a max pool operation over its word embeddings: given the short length of T3 codes, this proved to work well and it is similar to findings in relation extraction and targeted sentiment analysis~\cite{tang2015effective}.
We replicate $e_{T3}$ $n$ times and concatenate it to the text's word embeddings $x$ (Figure~\ref{fig:architecture}).

\noindent
\textit{Encoding Layer. }
We obtain a hidden representation $\vec{h}_{text}$
with a forward LSTM~\cite{gers1999learning}
over the concatenated input.
We then apply attention to capture the key parts of the input text w.r.t.~the given T3.
In detail, given the output matrix of the LSTM layer $H = [h_1, ..., h_n]$, we produce a hidden representation 
$h_{text}$ as follows:

{\centering
  $ \displaystyle
  \begin{aligned}
M&=tanh(
\begin{bmatrix}
W_h H\\
W_v e_{upv} \otimes e_N
\end{bmatrix}
)\\
\alpha_{text}&=softmax(w^TM)\\
h_{text}&=H\alpha^T
\end{aligned}
  $ 
\par}

\noindent
This is similar in principle to the attention-based LSTM by~\newcite{wang2016attention}, and proved to work better than classic attention over $H$ on our data.

\noindent
\textit{Decoding Layer. }
We predict $\hat{y} \in [0,1]$ with a dense layer followed by a sigmoidal activation.

\subsubsection{Including Description Information}
Each T3 comes with a short description, which was written by domain experts and used during manual labelling 
(the complete list is in the Appendix A).
We integrate information from such descriptions into our model as follows:
given the ordered word embeddings from the UPV description $(e_1, ..., e_d)$, 
we obtain a description representation $h_{descr}$ following the same steps as for the sample text.

In line with previous studies on siamese networks~\cite{yan2018few}, we observe better results when sharing the weights between the two LSTMs.
We keep two separated attention layers for sample texts and descriptions.
We concatenate $h_{text}$ and $h_{descr}$ and feed the obtained vector to the output layer.


\subsubsection{Multi-task Training}
\label{multitask_training}
A clear hierarchy exists between T3, T2 and T1 values (Section~\ref{sec:upv_theory}).
We integrate such information
using multi-task learning~\cite{caruana1997multitask,DBLP:journals/corr/Ruder17a}.
Given an input sample, we predict 
its relatedness not only w.r.t.~a T3 label, but also with 
its corresponding T2  and T1 labels\footnote{The mapping between sample and correct labels [T3, T2, T1] is as follows: \textit{positive}: [1, 1, 1];
\textit{slightly negative}: [0, 1, 1];
\textit{negative}: [0, 0, 1];
\textit{strictly negative}: [0, 0, 0].}.
In practice, given 
the hidden representation $h = h_{text} \oplus h_{descr}$,
we first
feed it into a dense layer $dense_{T1}$ to obtain $h_{T1}$, and predict 
$\hat{y}_{T1}$ with a sigmoidal function.
We then concatenate $h_{T1}$ with the previously obtained $h$,
and we predict $\hat{y}_{T2}$ with a T2-specific dense layer $\sigma(dense_{T2}(h \oplus h_{T1}))$.
Finally, $\hat{y}_{T3}$ is predicted as $\sigma(dense_{T3}(h \oplus h_{T2}))$.

In this way, the prediction $\hat{y}_i$ is based on both the original $h$ and the hidden representation computed in the previous stage of the hierarchy, $h_{i-1}$
(Figure~\ref{fig:architecture}).
%

\begin{table}[t]
\small\centering
\begin{tabular}{l
P{1.2cm}P{1.2cm}P{1.2cm}P{1.2cm}}
    \toprule
    &text & +att & +descr & +att+descr\\
    \midrule
    P &  77.5 & 78.1& \textbf{80.4} & 78.9\\
    R &  65.5 & \textbf{71.0}& 66.5 & 70.6\\
    $F_1$ & 71.0 & 74.2& 72.8 & \textbf{74.4}\\
    \bottomrule
    \end{tabular}
    \caption{Results of ablation study (single-task).}%
  \label{tab:results_models}%
  \end{table}

%
%

\section{Experiments and Discussion}

\subsection{Experimental Setting}
\label{sec:experimental_setting}
\subsubsection{Data Preparation}

For each positive sample, we generate 40 negative samples
%
(we found empirically that this was the best performing ratio, see Appendix C).

Moreover, 
to expose the system to more diverse input,
we slightly deform
the sample's text
when generating negative samples.
Following~\newcite{wei2019eda}, we implement 4 operations: random deletion, swap, insertion, and semantically-motivated substitution.
We also implement character swapping to increase the system's robustness to spelling errors (Figure~\ref{fig:augm}).

We consider only samples belonging to UPV labels with a support higher than 30 in the S2I corpus, thus rejecting 12 very rare UPVs. 
We select a random 80\% proportion from the data as training set; out of the remaining 980 samples,
we randomly select 450 as dev and use the rest as test set.
%
%

\subsubsection{Training Setting}

In order to allow for robust handling of OOVs, typos and spelling errors in the data,
we use FastText subword-informed pretrained vectors~\cite{bojanowski2017enriching} to initialise the word embedding matrix. 
%
%
%
We train using binary cross-entropy loss, with early stopping monitoring the development set loss with a patience of 5.
Sample weighting was used to account for the different error seriousness (1 for \textit{negative} and \textit{strictly neg} and 0.5 for \textit{mildly neg}).
Network hyperparameters are reported in Appendix C for replication.

\begin{table}[t]
    \centering
    \small
    \begin{tabular}{p{3mm}p{3mm}
    P{15mm}P{15mm}P{15mm}}
    \toprule
    &&\multicolumn{3}{c}{Multi-task train setting}\\
    \cmidrule{3-5}
    \multicolumn{2}{c}{Label} &
    T3 &
    T2+T3 &
    T1+T2+T3 \\
    \midrule
    \multirow{3}{*}{T3} &
    P       & 78.9 & \textbf{83.5} & 79.5 \\
    & R     & 70.6 & 67.0 & \textbf{72.0} \\
    & $F_1$ & 74.4 & 74.4 & \textbf{75.4} \\
    \midrule 
    \multirow{3}{*}{T2} &
    P &       -- & \textbf{92.0} & 84.9 \\
    & R &     -- & 40.5 & \textbf{62.3} \\
    & $F_1$ & -- & 56.2 & \textbf{71.9} \\
    \midrule
     \multirow{3}{*}{T1} &
    P       & -- & -- & 89.8   \\
    & R     & -- & -- & 70.1 \\
    & $F_1$ & -- & -- & 78.7 \\
    \bottomrule
    \end{tabular}
    \caption{Results 
    considering all granularities 
    and all (multi-)task training settings (T3, T2+T3, T1+T2+T3).}
    \label{tab:results}
\end{table}


\subsection{Results and Discussion}
\label{sec:results}

\begin{table}[t!]
    \centering
    \footnotesize
    \begin{tabular}{p{.005\textwidth} 
                    p{.12\textwidth}  
                    P{.03\textwidth} 
                    P{.03\textwidth} 
                    P{.03\textwidth} 
                    P{.03\textwidth} 
                    P{.03\textwidth} }
    \toprule
    T1 & T3 & P & R & $F_1$ &\multicolumn{2}{c}{Support (\%)} \\
    \cmidrule{1-7}
    \multirow{8}{*}{\begin{sideways}{\color{Fuchsia}\textit{\textbf{Emotional}}}\end{sideways}}
 & Harmony & 16.7 & 50.0 & 25.0 & 47 & 0.9\\
 & Appealing & 30.0 & 75.0 & 42.9 & 85 & 1.7\\
 & Aesthetics & 08.8 & 60.0 & 15.4 & 45 & 0.9\\
 & Comfort & 52.0 & 52.0 & 52.0 & 226 & 4.4\\
 & Entertainment & 40.0 & 54.5 & 46.2 & 108 & 2.1\\
 & Memorability & 16.7 & 12.5 & 14.3 & 77 & 1.5\\
 & Safety & 59.4 & 76.0 & 66.7 & 233 & 4.6\\
 & Sec.~People & 46.2 & 75.0 & 57.1 & 113 & 2.2\\
 \cmidrule{1-7}
 \multirow{2}{*}{\begin{sideways}{\color{PineGreen}\textit{\textbf{Epist}}}\end{sideways}}
 & Info.~Access & 84.6 & 55.0 & 66.7 & 198 & 3.9\\
 & Knowl.~attain. & 06.2 & 09.8 & 07.5 & 433 & 8.5\\
 \cmidrule{1-7}
 \multirow{15}{*}{\begin{sideways}{\color{RubineRed}\textit{\textbf{Function}}}\end{sideways}} 
 & Communication & 05.4 & 58.8 & 10.0 & 156 & 3.1\\
 & Mobile Acc. & 81.8 & 81.8 & 81.8 & 54 & 1.1\\
 & Mobility & 79.4 & 81.8 & 80.6 & 466 & 9.1\\
 & Multipurpose & 57.1 & 33.3 & 42.1 & 111 & 2.2\\
 & Availability & 01.4 & 33.3 & 02.6 & 104 & 2.0\\
 & Time Benefit & 51.9 & 66.7 & 58.3 & 217 & 4.3\\
 & Time Manag. & 76.9 & 83.3 & 80.0 & 102 & 2.0\\
 & Unburden & 41.9 & 72.0 & 52.9 & 190 & 3.7\\
 & Cap.~Expend. & 85.0 & 53.1 & 65.4 & 241 & 4.7\\
 & School Fees & 94.4 & 73.9 & 82.9 & 240 & 4.7\\
 & Econ.~Oppor. & 80.4 & 86.3 & 83.2 & 957 & 18.8\\
 & Effectiveness & 17.1 & 24.0 & 20.0 & 157 & 3.1\\
 & Lastingness & 83.3 & 38.5 & 52.6 & 116 & 2.3\\
 & Productivity & 52.4 & 66.7 & 58.7 & 200 & 3.9\\
 & Usability & 25.0 & 33.3 & 28.6 & 75 & 1.5\\
 \cmidrule{1-7}
 \multirow{5}{*}{\begin{sideways}{\color{TealBlue}\textit{\textbf{Indigen.}}}\end{sideways}}
 & Celebration & 100 & 50.0 & 66.7 & 55 & 1.1\\
 & Manners & 83.3 & 45.5 & 58.8 & 100 & 2.0\\
 & Morality & 20.0 & 22.2 & 21.1 & 98 & 1.9\\
 & Tradition & 85.7 & 70.6 & 77.4 & 175 & 3.4\\
 & Faith & 96.7 & 96.7 & 96.7 & 245 & 4.8\\
 \cmidrule{1-7}
 \multirow{10}{*}{\begin{sideways}\color{BurntOrange}\textit{\textbf{Intrinsic Human}}\end{sideways}} 
 & Longevity & 09.1 & 60.0 & 15.8 & 46 & 0.9\\
 & Healthc.~Acc. & 72.2 & 76.5 & 74.3 & 176 & 3.4\\
 & Treatment & 78.3 & 85.7 & 81.8 & 218 & 4.3\\
 & Educ.~Acc. & 80.0 & 54.5 & 64.9 & 103 & 2.0\\
 & Energy Acc. & 82.1 & 84.2 & 83.1 & 280 & 5.5\\
 & Food Security & 64.9 & 87.7 & 74.6 & 519 & 10.2\\
 & Shelter & 42.9 & 54.5 & 48.0 & 92 & 1.8\\
 & Water Access & 68.2 & 78.9 & 73.2 & 158 & 3.1\\
 & Water Quality & 37.0 & 90.9 & 52.6 & 148 & 2.9\\
 & Wellbeing & 09.8 & 59.1 & 16.9 & 245 & 4.8\\
 \cmidrule{1-7}
 \multirow{9}{*}{\begin{sideways}\color{RedOrange}\textit{\textbf{Social Significance}}\end{sideways}}
 & Appearance & 62.5 & 71.4 & 66.7 & 88 & 1.7\\
 & Dignity & 85.7 & 60.0 & 70.6 & 123 & 2.4\\
 & Pers.~Perf. & 33.3 & 11.1 & 16.7 & 111 & 2.2\\
 & Aspiration & 56.2 & 56.2 & 56.2 & 186 & 3.6\\
 & Modernisation & 57.1 & 40.0 & 47.1 & 98 & 1.9\\
 & Reputation & 52.9 & 69.2 & 60.0 & 189 & 3.7\\
 & Fam.~Caring & 63.6 & 58.3 & 60.9 & 258 & 5.1\\
 & Role Fulf. & 37.5 & 50.0 & 42.9 & 126 & 2.5\\
 & Togetherness & 53.3 & 57.1 & 55.2 & 132 & 2.6\\
 \midrule
 & \textit{\textbf{Total}} & 
 \textit{44.9} & \textit{70.3} & \textit{50.5} \\
 \bottomrule
    \end{tabular}
    \caption{
    Single label results in the \textit{Real-World Simulation} setting, with label support in S2I corpus.
    }
    \label{tab:single_results}
\end{table}

\subsubsection{Models Performance}
During experiments, we monitor precision, recall and $F_1$ score.
For evaluation, we consider a test set where negative samples appear \textit{in the same proportion} as in the train set (1/40 positive/negative ratio).
The results of our experiments are reported in Table~\ref{tab:results_models}.
Notably, adding attention and integrating signal from descriptions to the base system lead to significant improvements in performance.
%

\subsubsection{Multi-task Training}
We consider the best performing model
and run experiments with the three considered multi-task train settings (Section~\ref{multitask_training}).
We consider 3 layers of performance, corresponding to
{T3}, {T2} and {T1} labels.
This is useful because, in the application context, different levels of granularity can be monitored.
As shown in Table~\ref{tab:results}, we observe
relevant improvements in F1 scores when jointly learning more than one training objective.
This holds true not only for T3 classification, but also for T2 classification when training with the T3+T2+T1 setting.
This seems to indicate that
the signal encoded in the additional training objectives
indirectly conveys information about the label hierarchy which is indeed useful for classification.

\subsubsection{Real-World Simulation and Error Analysis}
To simulate a real scenario where we annotate a new interview with the corresponding UPVs,
we perform further experiments on the test set
by generating, for each sample, \textit{all possible} negative samples.
%
We annotate using the T1+T2+T3 model, finetuning the threshold for each UPV on the development set,
and perform a detailed error analysis of the results on the test set.
%

As reported in Table~\ref{tab:single_results},
we observe a significant drop in precision,
which confirms the extreme difficulty of the task in a real-world setting due to the extreme data imbalancy. 
Note, however, that
recall remains relatively stable over changes in evaluation
settings.
This is particularly important for a system which is meant to enhance the annotators' speed,
rather than to completely replace human experts:
in this context, missing labels are more time consuming to recover than correcting false positives.

Not surprisingly, particularly good performance is often obtained on T3 labels which tend to correlate with specific terms (as \textit{School Fees}, or \textit{Faith}).
In particular, we observe a correlation between a T3 label's support in the corpus and the system's precision in predicting that label:
with very few exceptions, all labels where the system obtained a precision lower than 30 had a support similar or lower than 3\%.

The analysis of the ROC curves shows that, overall, satisfactory results are obtained
for all T1 labels considered (Appendix D), leaving, however, considerable room for future research.

\section{Conclusions and Future Work}

In this study, we provided a first stepping stone towards future research at the intersection of NLP and Sustainable Development (SD).
As a case study, we investigated the opportunity of NLP to enhancing project sustainability
through improved community profiling
by providing a cost effective way towards structuring qualitative data.

This research is in line with a general call for AI towards social good, where the potential positive impact of NLP is notably missing.
In this context, we proposed the new challenging task of \textit{Automatic User-Perceived Values Classification}: we provided the task definition, an annotated dataset (the \textit{Stories2Insights} corpus) and a set of light (in terms of overall number of parameters) neural baselines for future reference. 

Future work will investigate ways to improve performance (and especially precision scores) on our data, in particular on low-support labels.
Possible research direction could include more sophisticated thresholding selection techniques~\cite{fan2007study,DBLP:journals/ml/ReadPHF11} to replace the simple threshold finetuning which is currently used for simplicity.
While 
deeper and computationally heavier models as~\newcite{DBLP:conf/naacl/DevlinCLT19} could possibly obtain notable gains in performance on our data,
it is the responsibility of the NLP community – especially with regards to social good applications –
to provide solutions which don't penalise countries suffering from access biases (as contexts with low access to computational power), as it is the case of many developing countries.

We hope our work will spark interest and 
open a constructive dialogue between the fields of NLP and SD,
and result in new interesting applications. 

\bibliography{ms}
\bibliographystyle{acl_natbib}

\clearpage

\onecolumn
\section*{Appendix A – Definitions of User-Perceived Values}
\label{appendix_a}
\small{
\begin{longtable}[c]{|p{.02\textwidth} | p{.24\textwidth} | p{.68\textwidth} |}
\cline{1-3}
\multirow{11}{*}{\begin{sideways}{\color{Fuchsia}\textit{\textbf{Emotional}}}\end{sideways}} &  
\multicolumn{2}{c|}{\textbf{{\color{Fuchsia}Conscience}}}\\
\cline{2-3}
\multicolumn{1}{|c|}{} &    Preservation of Environment & 	Preservation of natural resources \\
\multicolumn{1}{|c|}{} &	Harmony & Being at peace with one another \\
\cline{2-3}
\multicolumn{1}{|c|}{} & \multicolumn{2}{c|}{\textbf{{\color{Fuchsia}Contentment}}}\\
\cline{2-3}
\multicolumn{1}{|c|}{} & Appealing Senses & 	Being pleasing to the senses taste and smell \\
\multicolumn{1}{|c|}{} & Aesthetics Items & 	Physical appearance of item or person which is pleasing to look at \\
\multicolumn{1}{|c|}{} &Comfort & 	State of being content, having a positive feeling \\
\multicolumn{1}{|c|}{} &Entertainment & 	Something affording pleasure, diversion or amusement \\
\multicolumn{1}{|c|}{} &Memorability & 	Association to a past event with emotional significance \\
\cline{2-3}
\multicolumn{1}{|c|}{}  &  \multicolumn{2}{c|}{\textbf{{\color{Fuchsia}Human Welfare}}}\\
\cline{2-3}
\multicolumn{1}{|c|}{} & Safety (Animals Items Nature) & 	Being protected from or prevent injuries or accidents by animals or nature \\
\multicolumn{1}{|c|}{} & Security People & 	Being free from danger and threat posed by people \\
\cline{1-3}
 \multirow{4}{*}{\begin{sideways}{\color{PineGreen}\textit{\textbf{Epistemic}}}\end{sideways}} &     \multicolumn{2}{c|}{\textbf{{\color{PineGreen}Information}}}\\
\cline{2-3}
   \multicolumn{1}{|c|}{} & Information Access & 	Ability to stay informed \\
\cline{2-3}
\multicolumn{1}{|c|}{} &\multicolumn{2}{c|}{\textbf{{\color{PineGreen}Knowledge}}}\\
\cline{2-3}
 \multicolumn{1}{|c|}{} &   Knowledge attainment & 	The ability to learn or being taught new knowledge \\
\cline{1-3}
\multirow{23}{*}{\begin{sideways}{\color{RubineRed}\textit{\textbf{Function}}}\end{sideways}} &  
\multicolumn{2}{c|}{\textbf{{\color{RubineRed}Convenience}}}\\
\cline{2-3}
 \multicolumn{1}{|c|}{} & 	Banking Access & 	Having continuous access to banking services \\
 \multicolumn{1}{|c|}{} & 	Communication & 	Ability to interact with someone who is far \\
 \multicolumn{1}{|c|}{} & 	Mobile Phone Access & 	Having continuous access to mobile telecommunication services \\
   \multicolumn{1}{|c|}{} &   Mobility & 	Being able to transport goods, or to carry people from one place to another \\
   \multicolumn{1}{|c|}{} &   Multipurpose & 	Able to be used for a multitude of purposes \\
   \multicolumn{1}{|c|}{} &   Portable & 	An item that can easily be carried, transported or conveyed by hand \\
   \multicolumn{1}{|c|}{} &   Availability & 	Possible to get, buy or find in the area \\
   \multicolumn{1}{|c|}{} &   Time Benefit & 	Accomplish something with the least waste of time or minimum expenditure of time \\
   \multicolumn{1}{|c|}{} &   Time Management & 	Being able to work or plan towards a schedule \\
   \multicolumn{1}{|c|}{} &   Unburden & 	Making a task easier by simplifying \\
   \cline{2-3}
\multicolumn{1}{|c|}{} &\multicolumn{2}{c|}{\textbf{{\color{RubineRed}Cost Economy}}}\\\cline{2-3}
  \multicolumn{1}{|c|}{} &  Capital Expenditure & 	Cost savings achieved \\
   \multicolumn{1}{|c|}{} & School Fees & 	Ability to pay for school fee \\
\cline{2-3}
\multicolumn{1}{|c|}{} & \multicolumn{2}{c|}{\textbf{{\color{RubineRed}Income Economy}}}\\
\cline{2-3}
   \multicolumn{1}{|c|}{} & Economic Opportunity & Obtaining cash, assets, income through one-off sales or ongoing business opportunities \\
   \multicolumn{1}{|c|}{} & Barter Trade & 	Non-monetary trade of goods or services \\
   \cline{2-3}
\multicolumn{1}{|c|}{} &\multicolumn{2}{c|}{\textbf{{\color{RubineRed}Quality and Performance}}}\\
\cline{2-3}
\multicolumn{1}{|c|}{} &Effectiveness & 	Adequate to accomplish a purpose or producing the result \\
  \multicolumn{1}{|c|}{} &  Lastingness & 	Continuing or enduring a long time \\
  \multicolumn{1}{|c|}{} &  Productivity & 	Rate of output and means that lead to increased productivity \\
\multicolumn{1}{|c|}{} &	Reliability & 	The ability to rely or depend on operation or function of an item or service \\
\multicolumn{1}{|c|}{} &	Usability & 	Refers to physical interaction with item being easy to operate handle or look after \\
\cline{1-3}
\multirow{7}{*}{\begin{sideways}{\color{TealBlue}\textit{\textbf{Indigenous}}}\end{sideways}} 
 &\multicolumn{2}{c|}{\textbf{{\color{TealBlue}Social Norm}}}\\
 \cline{2-3}
\multicolumn{1}{|c|}{} &	Celebration & 	Association chosen as they play important part during celebration \\
\multicolumn{1}{|c|}{} &	Manners & 	Ways of behaving with reference to polite standards and social components \\
\multicolumn{1}{|c|}{} &	Morality & 	Following rules and the conduct \\
\multicolumn{1}{|c|}{} &	Tradition & 	Expected form of behaviour embedded into the specific culture of city or village \\\cline{2-3}
\multicolumn{1}{|c|}{} &\multicolumn{2}{c|}{\textbf{{\color{TealBlue}Religion}}}\\\cline{2-3}
\multicolumn{1}{|c|}{} &	Faith & 	Belief in god or in the doctrines or teachings of religion \\
	\cline{1-3}
\multirow{14}{*}{\begin{sideways}\color{BurntOrange}\textit{\textbf{Intrinsic Human}}\end{sideways}}  &	
\multicolumn{2}{c|}{\textbf{{\color{BurntOrange}Health}}}\\
\cline{2-3}
\multicolumn{1}{|c|}{} &	Longevity & 	Means that lead to an extended life span \\
\multicolumn{1}{|c|}{} &	Health Care Access & 	Being able to access medical services or medicine \\
\multicolumn{1}{|c|}{} &	Treatment & 	To require a hospital or medical attention as a consequence of illness or injury \\
\multicolumn{1}{|c|}{} &	Preserv. of Health & 	Practices performed for the preservation of health \\
\cline{2-3}
\multicolumn{1}{|c|}{} &
\multicolumn{2}{c|}{\textbf{{\color{BurntOrange}Physiological}}}\\\cline{2-3}
\multicolumn{1}{|c|}{} &	Education Access & 	Being able to access educational services \\
\multicolumn{1}{|c|}{} &	Energy Access & 	Being able to obtain energy services or resources \\
\multicolumn{1}{|c|}{} &			Food Security & 	The ability to have a reliable and continuous supply of food \\
\multicolumn{1}{|c|}{} &			Shelter & 	A place giving protection from bad weather or danger \\
\multicolumn{1}{|c|}{} &			Water Access & 	Continuous access or availability of water \\
\multicolumn{1}{|c|}{} &			Water Quality & 	To have clean water as sickness, colour and taste \\\cline{2-3}
\cline{2-3}
\multicolumn{1}{|c|}{} &\multicolumn{2}{c|}{\textbf{{\color{BurntOrange}Quality of Life}}}\\\cline{2-3}
\multicolumn{1}{|c|}{} &			Wellbeing & 	Obtaining good or satisfying living condition (for people or for the community) \\
\cline{1-3}
\multirow{6}{*}{\begin{sideways}\color{RedOrange}{
\textbf{\textit{Significance}}}\end{sideways}}  &\multicolumn{2}{c|}{\textbf{{\color{RedOrange}Identity}}}\\\cline{2-3}
	\multicolumn{1}{|c|}{} &	Appearance & 	Act or fact of appearing as to the eye or mind of the public \\
	\multicolumn{1}{|c|}{} &	Belongingness & 	Association with a certain group, their values and interests \\
	\multicolumn{1}{|c|}{} &	Dignity & 	The State or quality of being worthy of honour or respect \\
	\multicolumn{1}{|c|}{} &	Personal Performance & 	The productivity to which someone executes or accomplishes work \\
		\cline{2-3}
		\multicolumn{1}{|c|}{}
		&\multicolumn{2}{c|}{\textbf{{\color{RedOrange}Status}}}\\\cline{2-3}
		\multicolumn{1}{|c|}{} &Aspiration & 	Desire or aim to become someone better or more powerful or wise \\
	\multirow{6}{*}{\begin{sideways}\color{RedOrange}{
\textbf{\textit{Social}}}\end{sideways}} &
	Modernisation & 	Transition to a modern society away from a traditional society \\
	\multicolumn{1}{|c|}{} &		Reputation & 	Commonly held opinion about ones character \\
		\cline{2-3}
		\multicolumn{1}{|c|}{} &\multicolumn{2}{c|}{\textbf{{\color{RedOrange}Social Interaction}}}\\\cline{2-3}
	\multicolumn{1}{|c|}{} &	Altruism & 	The principle and practice of unselfish concern \\
	\multicolumn{1}{|c|}{} &		Family Caring & 	Displaying kindness and concern for family members \\
	\multicolumn{1}{|c|}{} &		Role Fulfilling & 	Duty to fulfilling tasks or responsibilities associated with a certain role \\
	\multicolumn{1}{|c|}{} &		Togetherness & 	Warm fellowship, as among friends or members of a family \\
	\cline{1-3}
\end{longtable}
}
\twocolumn

\begin{figure*}[h!]
    \centering
    \includegraphics[width=\textwidth]{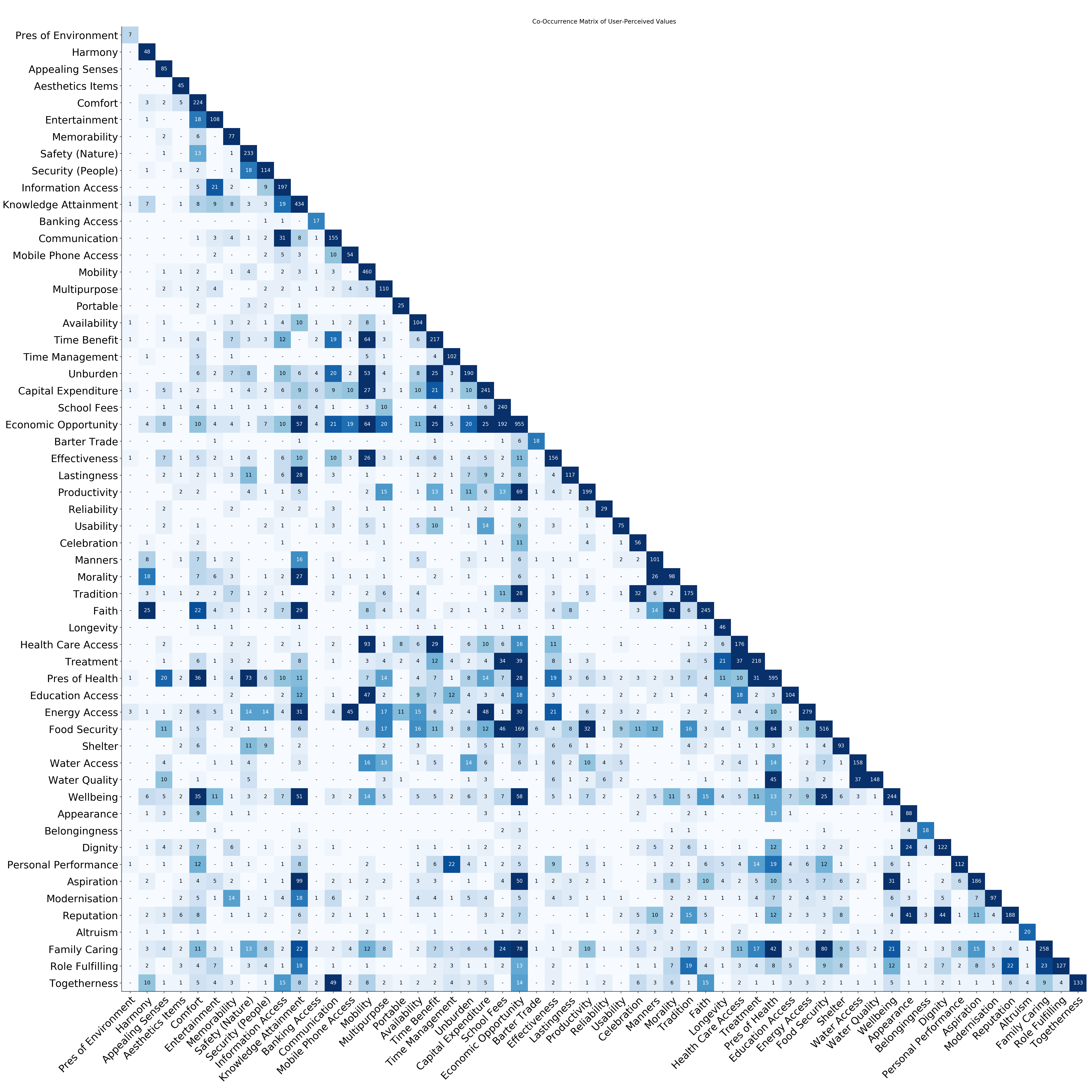}
    \caption{Co-occurrence matrix of T3 labels in the S2I corpus.}
    \label{fig:co_occurrence_matrix}
\end{figure*}


\section*{Appendix B – Co-occurrence matrix of User-Perceived Values in the S2I corpus.}
\normalsize
The co-occurrence matrix in Figure~\ref{fig:co_occurrence_matrix} depicts the inter-relatedness between different T3 labels.
The intensity of colour corresponds to the number of samples in the S2I corpus where the given T3 labels co-occur.

The analysis of labels co-occurrence can offer valuable insights on commonly associated User-Perceived Values (UPVs, \cite{HIRMER2014145}):
this can be useful to highlight challenges and problems in the considered community, which might not be known to the dwellers themselves. While some correlations are typical and expected, others are related to the specific Ugandan context, and might be surprising to those external to the location.

For example, \textit{Economic Opportunity}, \textit{Food Security} and \textit{Preservation of Health} appear to frequently co-occur with other T3 labels.
Note that the lack of employment opportunity, the availability of food and the quality of healthcare services represent endemic problems in the rural context studied in this paper.
As they constitute primary concerns 
for most interviewees, it is therefore unsurprising that they were mentioned frequently in relation to many of the items selected as part of the UPV game (Section 4).
%
%
A further illustrative example of the cultural context - in this case rural Uganda - is the high concurrence of \textit{Unburden} and \textit{Mobility}. This can be explained by the fact that rural roads are often of poor quality and villages or areas are inaccessible by motorised vehicles. Henceforth, people are required to find alternatives moves of transport for moving themselves to hospital or crops to the nearest market for sell, for example.
As a final example, the frequent mentioning of \textit{Faith}, \textit{Harmony} and \textit{Morality}, which also tend to co-occur in similar contexts, testifies the fundamental role played by religion in the rural villages considered in this study.

The information on the {(co-)}occurrence of UPVs in a community is also particularly valuable for designing appropriate project communication (Figure 1b),
which can increase project buy-in through focused messaging (Section 3).




\section*{Appendix C – Experimental Specifications.}
\normalsize

In this Appendix, we report on the exact experimental setting used for experiments
to aid experiment reproducibility. 

\subsection*{C.1~Data Specifications}

\begin{figure}[h!]
    \centering
    \includegraphics[width=\columnwidth]{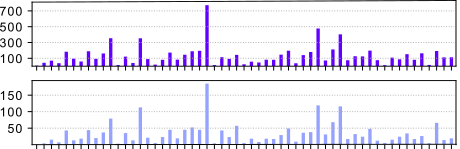}
    \caption{From top to bottom: distribution of UPVs in the training and heldout (dev+test) sets. Labels in the $x$ axis follow the same order as in Figure~3 of the main document.}
    \label{fig:train_dev_distr}
\end{figure}

\vspace{1mm}
\noindent
\textbf{Data Selection and Splitting.~}
We select all sentences from the 119 interviews which were at least 3 tokens long and which were annotated with at least one UPV.
We then randomly select an 80\% proportion of the data as training set, and take the rest as heldout data (with a dev/test split of respectively 450 and 530 samples).
Figure~\ref{fig:train_dev_distr} shows that the obtained label distribution is similar.

\vspace{1mm}
\noindent
\textbf{Data Anonymization.~}
In order to prevent the identification of the interviewees~\cite{sweeney2000simple},
data was manually anonymized. We anonymized all occurrences of: proper names, names of villages, cities or other geographical elements, and other names that might be sensitive (as names of tribes, languages, ...).

\vspace{1mm}
\noindent
\textbf{Data Sample.~}
We are providing a sample of the data in the supplementary material.
Each data sample is associated with the following fields:
\begin{itemize}[noitemsep]
    \item \textit{id}: a unique identifier of the sample; 
    \item \textit{text}: a sentence to be classified;
    \item \textit{t3\_labels}: a list of the gold T3 labels associated with the sample.
\end{itemize}

\noindent
For privacy reasons, we are not releasing metadata information associated with the samples (as the interviewee's name, gender, age, or the exact village name).

\vspace{1mm}
\noindent
\textbf{Data Preprocessing.~}
For sentence splitting and word tokenization, we used NLTK's
\texttt{sent\_tokenize} and \texttt{word\_tokenize}
tokenizers~\cite{loper2002nltk}\footnote{\url{https://www.nltk.org/api/nltk.tokenize.html}}.
We use a set of regex to find interviewer comments and questions. Given that \textit{Why-Probing} (Section~4.1, \citet{reynolds2001laddering}) was used, interviewers' comments are very limited and standard.

\vspace{1mm}
\noindent
\textbf{Negative Samples Generation.~}
To generate negative samples (Section~6.1),
we slightly modify \newcite{wei2019eda}'s  implementation\footnote{\url{https://github.com/jasonwei20/eda_nlp/blob/d75e8bd4631f4d93260cb291aa47852d8eacd51d/code/eda.py\#L65}}
EDA (Easy Data Augmentation techniques)
by adding a new function for character swapping and by adapting the stopword list.
For semantic-based replacement, 
we rely on NLTK's interface\footnote{\url{https://www.nltk.org/_modules/nltk/corpus/reader/wordnet.html}} to
WordNet~\cite{fellbaum2012wordnet}.
Random shuffling and choice are controlled by a seed.

\subsection*{C.2~Further Specifications}

\textbf{(Hyper)-Parameters Selection.~}
All parameters used for experiments are reported in Table~\ref{tab:hyperpars}.

\begin{table}[h!]
\centering\small
    \begin{tabular}{
    lrlr
    }
    \toprule
    \multicolumn{2}{l}{parameter \hfill value}&
    \multicolumn{2}{l}{parameter \hfill value}\\
    \cmidrule(lr){1-2}
    \cmidrule(lr){3-4}
    \textit{mildly neg} s.~ratio & 5 & embedding size & 300\\
    \textit{neg} sample ratio & 11
    &  LSTM hid.~size & 128\\
    \textit{strictly neg} s.~ratio & 24
    &  dropout (all l.) & 0.2\\
    max sample len & 25
    &  batch size & 32 \\
    max descr len & 15 
    & no epochs & 70 \\
    max UPV code len & 4 &
    \multicolumn{2}{l}{optimizer \hfill\textit{Adam}}\\
    \bottomrule
    \end{tabular}
  \caption{Adopted (hyper-)parameters.}%
  \label{tab:hyperpars}
\end{table}

\noindent
We use 300-dimensional FastText subword-informed
pretrained vectors~\cite{bojanowski2017enriching}\footnote{We chose the \texttt{wiki.en.zip} model pretrained on the English Wikipedia \url{https://fasttext.cc/docs/en/pretrained-vectors.html}}
to get the word embedding representations for each input sample.

Note that the goal of this paper is to present a new interesting NLP application, namely NLP for Sustainable Development:
therefore, our goal here is to provide a set of robust baselines on our new S2I dataset, which can be referenced for future research.
For this reason, we don't perform extensive hyper-parameter tuning on the selected models.

The only parameters we optimize are the number of generated negative samples of each type (\textit{mildly negative, negative} and \textit{strictly negative}).
The best ratios were found empirically through experiments.
The ratio used for optimization are reported in Table~\ref{tab:neg_sample_ratio}.

\begin{table}[h!]
    \centering\small
    \begin{tabular}{cccc}
    \toprule
    total
    & mildly negative
    & negative
    & strictly negative
    \\
    \midrule
    0  & 0 & 0 & 0 \\
    5  & 1 & 2 & 2 \\
    10 & 2 & 2 & 6 \\
    15 & 3 & 4 & 8 \\
    20 & 4 & 7 & 9 \\
    25 & 5 & 8 & 12 \\
    30 &  5 & 11 & 14 \\
    35 &  5 & 11 & 19 \\
    \textbf{40} &  
    \textbf{5 } & 
    \textbf{11} & 
    \textbf{24} \\
    45 &  5 & 10 & 30 \\
    50 &  5 & 12 & 33 \\
    55 &  5 & 13 & 37 \\
    60 &  5 & 14 & 41 \\
    \bottomrule
    \end{tabular}
    \caption{Details of the relative number of \textit{mildly negative, negative} and \textit{strictly negative} samples used for experiments. Best ratio (used in all reported experiments) is in bold.}
    \label{tab:neg_sample_ratio}
\end{table}

\begin{figure}[h!]
    \centering
    \includegraphics[width=0.9\columnwidth]{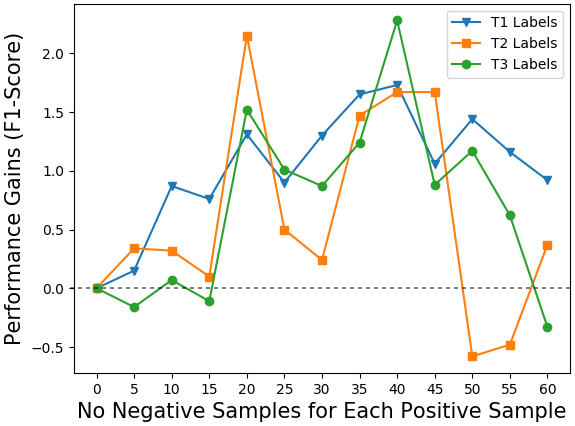}
    \caption{Progression of performance gains in F1-Score, considering the three labels T1, T2 and T3.}
    \label{fig:ratios_training}
\end{figure}

\noindent
The analysis of the performance progression over training (Figure~\ref{fig:ratios_training}) shows that, in line with \newcite{wei2019eda}, adding negative examples is useful to improve performance:
in our case, the plateau is reached around 40 augmented samples.
In particular, we observe gains in all considered output levels (T1, T2 and T3 labels).

\vspace{1mm}
\noindent
\textbf{Number of Parameters and Runtime Specifications.~}
Table~\ref{tab:hyperpar_number} reports on the total number of (trainable) parameters and the average runtime/step for each considered model.
Embeddings are kept fixed over training to avoid overfitting.

\begin{table}[h!]
    \centering\small
    \begin{tabular}{@{}rlcc@{}}
    \toprule
    (Multi-)task & Model & \#pars & avg runtime/ \\
    Setting & & & step\\
    \midrule
    \multirow{4}{*}{T3}
    & text             & 373,377 & 55s\\
    & +att             & 590,013 & 56s\\
    & +descr           & 373,505 & 74s\\
    & +att+descr       & 806,777 & 75s\\
    \midrule
    T2T3  & +att+descr & 844,154 & 78s\\
    T1T2T3& +att+descr & 865,019 & 85s\\
    \bottomrule
    \end{tabular}
    \caption{Number of trainable parameters and average runtime/step for all considered models and (multitask) training settings.}
    \label{tab:hyperpar_number}
\end{table}

\vspace{1mm}
\noindent
\textbf{Computing Infrastructure.~}
We run experiments on an NVIDIA GeForce GTX 1080 GPU.

\vspace{1mm}
\noindent
\textbf{Evaluation Specifications.~}
For computing the evaluation metrics, we use the sklearn's~\cite{scikit-learn} implementation of precision, recall and $F_1$ score\footnote{\url{https://scikit-learn.org/stable/modules/classes.html\#module-sklearn.metrics}}.

\begin{figure*}[h!]
    \centering
    \includegraphics[width=\textwidth]{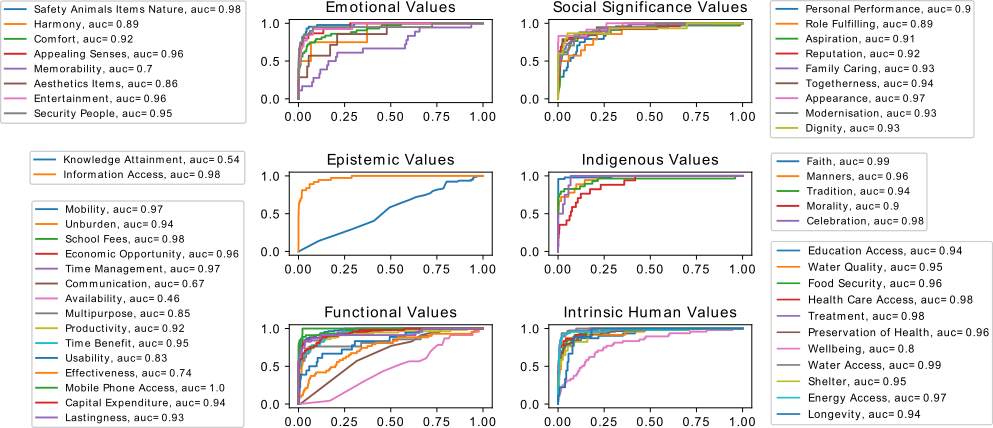}
    \caption{ROC curves for each T3 label, grouped by T1 categories.}
    \label{fig:roc_curves}
\end{figure*}

\newpage

\section*{Appendix D – Single-Label Performance.}

\normalsize
In this Appendix, we report the ROC curves for each T3 label, grouped by T1 categories.
Figure~\ref{fig:roc_curves} reports results obtained with the best performing model (Base+Attention+Description) trained with the T1+T2+T3 multi-task framework.

We evaluate with the ``real-world'' evaluation setting (Section 6.1),
that is, we generate \textit{all} positive and negative instances for each training sample.
In practice, for a test sample $x$ associated with the T3 labels $T3_2$ and $T3_{45}$, we would generate 50 test instances $\{(x, T3_1) \rightarrow 0, (x, T3_1) \rightarrow 0, (x, T3_2) \rightarrow 1, ..., (x, T3_{50}) \rightarrow 0 \}$, one for each of the T3 considered during training. All generated test samples would be negative, with the exception of $(x, T3_2)$ and $(x, T3_{45})$.

The single T3 labels' AUC show that satisfactory results are obtained overall for all T1 macro-labels:
in particular, we obtain an AUC $>= 70$ for 47 out of 50 labels.
Despite these promising results, our best model still struggles with some T3 labels, notably \textit{Knowledge Attainment}, \textit{Availability} and \textit{Communication}.
While the paper leaves ample room for future research,
preliminary results are encouraging.

Refer to Section 6.2 for further details.

\end{document}